\documentstyle[conll,11pt]{article}
\title{{\bf Introduction to the CoNLL-2000 Shared Task: Chunking}}
\author{
Erik F. Tjong Kim Sang \\ CNTS -- Language Technology Group \\
   University of Antwerp \\ \texttt{erikt@uia.ua.ac.be} 
\And
Sabine Buchholz \\ ILK, Computational Linguistics \\
   Tilburg University \\ \texttt{s.buchholz@kub.nl}
}

\date{\today}

\begin{document}

\maketitle

\setcounter{page}{127}
{
\begin{picture}(0,0)
\put(0,145){
\makebox(440,0){
In:
{\em Proceedings of CoNLL-2000 and LLL-2000},
pages 127--132,
Lisbon, Portugal, 2000.
}}
\end{picture}
}

\vspace*{-0.0cm}
\begin{abstract}
\noindent
We describe the CoNLL-2000 shared task: dividing text into
syntactically related non-overlapping groups of words, so-called 
text chunking.
We give background information on the data sets, present a general
overview of the systems that have taken part in the shared task and
briefly discuss their performance.
\end{abstract}

\section{Introduction}

Text chunking is a useful preprocessing step for parsing.
There has been a large interest in recognizing non-overlapping noun
phrases (Ramshaw and Marcus \shortcite{ramshaw95} and follow-up
papers) but relatively little has been 
written about identifying phrases of other syntactic categories.
The CoNLL-2000 shared task attempts to fill this gap.

\section{Task description}
\label{sec-task}

Text chunking consists of dividing a text into phrases in such a way 
that syntactically related words become member of the same phrase.
These phrases are non-overlapping which means that one word can only 
be a member of one chunk.
Here is an example sentence:

\begin{quote}
$[_{\rm NP}$ He $]$ 
$[_{\rm VP}$ reckons $]$ 
$[_{\rm NP}$ the current account deficit $]$ 
$[_{\rm VP}$ will narrow $]$
\\
$[_{\rm PP}$ to $]$ 
$[_{\rm NP}$ only $\pounds$ 1.8 billion $]$ 
\\
$[_{\rm PP}$ in $]$ 
$[_{\rm NP}$ September $]$ 
.
\end{quote}

\noindent
Chunks have been represented as groups of words between square
brackets.
A tag next to the open bracket denotes the type of the chunk.
As far as we know, there are no annotated corpora available which
contain specific information about dividing sentences into chunks of
words of arbitrary types.
We have chosen to work with a corpus with parse information, the Wall
Street Journal (WSJ) part of the Penn Treebank II corpus \cite{marcus93}, 
and to extract chunk information from the parse trees in this corpus.
We will give a global description of the various chunk types in the
next section.

\section{Chunk Types}

The chunk types are based on the syntactic category part (i.e.\ without function tag) of the bracket label in the Treebank (cf. Bies \shortcite{bies1995} p.35). Roughly, a chunk contains everything to the left of and including the syntactic head of the constituent of the same name. Some Treebank constituents do not have related chunks. The head of S (simple declarative clause) for example is normally thought to be the verb, but as the verb is already part of the VP chunk, no S chunk exists in our example sentence.

Besides the head, a chunk also contains premodifiers (like determiners and adjectives in NPs), but no postmodifiers or arguments. This is why the PP chunk only contains the preposition, and not the argument NP, and the SBAR chunk consists of only the complementizer.

There are several difficulties when converting trees into chunks. In the most simple case, a chunk is just a syntactic constituent without any further embedded constituents, like the NPs in our examples. In some cases, the chunk contains only what is left after other chunks have been removed from the constituent, cf. ``(VP loves (NP Mary))'' above, or ADJPs and PPs below. We will discuss some special cases during the following description of the individual chunk types.

\subsection{NP}
Our NP chunks are very similar to the ones of Ramshaw and Marcus
\shortcite{ramshaw95}. 
Specifically, possessive NP constructions are split in front of the possessive marker (e.g. $[_{\rm NP}$~Eastern Airlines~] $[_{\rm NP}$~' creditors~]) and the handling of coordinated NPs follows the Treebank annotators.
However, as Ramshaw and Marcus do not describe the details of their conversion algorithm, results may differ in difficult cases, e.g. involving NAC and NX.\footnote{E.g. (NP-SBJ (NP Robin Leigh-Pemberton) , (NP (NAC Bank (PP of (NP England))) governor) ,)
which we convert to
$[_{\rm NP}$~Robin Leigh-Pemberton~] , Bank $[_{\rm PP}$~of~] $[_{\rm NP}$~England~] $[_{\rm NP}$~governor~]
whereas Ramshaw and Marcus state that ` ``governor'' is not included in any baseNP chunk'.}

An ADJP constituent inside an NP constituent becomes part of the NP chunk:
\begin{quote}
(NP The (ADJP most volatile) form)\\ 
$\rightarrow$ $[_{\rm NP}$~the most volatile form~]
\end{quote}

\subsection{VP}
In the Treebank, verb phrases are highly embedded; see e.g. the following sentence which contains four VP constituents. Following Ramshaw and Marcus' V-type chunks, this sentence will only contain one VP chunk:
\begin{quote}
((S (NP-SBJ-3 Mr. Icahn) (VP may not (VP want (S (NP-SBJ *-3) (VP to (VP sell ...))))) . )) \\
$\rightarrow$ $[_{\rm NP}$~Mr. Icahn~] $[_{\rm VP}$~may not want to sell~] ... 
\end{quote}

\noindent
It is still possible however to have one VP chunk directly follow another: $[_{\rm NP}$~The impression~] $[_{\rm NP}$~I~] {\bf $[_{\rm VP}$~have got~] $[_{\rm VP}$~is~]} $[_{\rm NP}$~they~] $[_{\rm VP}$~'d love to do~] $[_{\rm PRT}$~away~] $[_{\rm PP}$~with~] $[_{\rm NP}$~it~]. 
In this case the two VP constituents did not overlap in the Treebank.

Adverbs/adverbial phrases become part of the VP chunk (as long as they are in front of the main verb):

\begin{quote}
(VP could (ADVP very well) (VP show ... ))\\
 $\rightarrow$ $[_{\rm VP}$~could very well show~] ...
\end{quote}
In contrast to Ramshaw and Marcus \shortcite{ramshaw95}, predicative adjectives of the verb are not part of the VP chunk, e.g. in ``$[_{\rm NP}$~they~] $[_{\rm VP}$~are~] $[_{\rm ADJP}$~unhappy~]''.

In inverted sentences, the auxiliary verb is not part of any verb phrase in the Treebank. Consequently it does not belong to any VP chunk:
\begin{quote}
((S (SINV (CONJP Not only) does (NP-SBJ-1 your product) (VP have (S (NP-SBJ *-1) (VP to (VP be (ADJP-PRD excellent)))))) , but ...\\
$\rightarrow$ $[_{\rm CONJP}$~Not only~] does $[_{\rm NP}$~your product~] $[_{\rm VP}$~have to be~] $[_{\rm ADJP}$~excellent~] , but ...
\end{quote}

\subsection{ADVP and ADJP}
ADVP chunks mostly correspond to ADVP constituents in the Treebank. However, ADVPs inside ADJPs or inside VPs if in front of the main verb are assimilated into the ADJP respectively VP chunk. On the other hand, ADVPs that contain an NP make two chunks:
\begin{quote}
(ADVP-TMP (NP a year) earlier) \\
$\rightarrow$ $[_{\rm NP}$~a year~] $[_{\rm ADVP}$~earlier~]
\end{quote}

\noindent
ADJPs inside NPs are assimilated into the NP.
And parallel to ADVPs, ADJPs that contain an NP make two chunks:
\begin{quote}
(ADJP-PRD (NP 68 years) old) \\
$\rightarrow$ $[_{\rm NP}$~68 years~] $[_{\rm ADJP}$~old~]
\end{quote}
It would be interesting to see how changing these decisions (as can be done in the Treebank-to-chunk conversion script\footnote{The Treebank-to-chunk conversion script is available from http://ilk.kub.nl/\~{ }sabine/chunklink/}) influences the chunking task.

\subsection{PP and SBAR}
Most PP chunks just consist of one word (the preposition) with the part-of-speech tag IN. This does not mean, though, that finding PP chunks is completely trivial. INs can also constitute an SBAR chunk (see below) and some PP chunks contain more than one word. This is the case with fixed multi-word prepositions such as {\em such as, because of, due to}, with prepositions preceded by a modifier: {\em well above, just after, even in, particularly among} or with coordinated prepositions: {\em inside and outside}.
We think that PPs behave sufficiently differently from NPs in a sentence for not wanting to group them into one class (as Ramshaw and Marcus did in their N-type chunks), and that on the other hand tagging all NP chunks inside a PP as I-PP would only confuse the chunker. We therefore chose not to handle the recognition of true PPs (prep.+NP) during this first chunking step.

SBAR chunks mostly consist of one word (the complementizer) with the part-of-speech tag IN, but like multi-word prepositions, there are also multi-word complementizers: {\em even though, so that, just as, even if, as if, only if}. 

\subsection{CONJP, PRT, INTJ, LST, UCP}
Conjunctions can consist of more than one word as well: {\em as well as, instead of, rather than, not only, but also}. One-word conjunctions (like {\em and, or}) are not annotated as CONJP in the Treebank, and are consequently no CONJP chunks in our data.

The Treebank uses the PRT constituent to annotate verb particles, and our PRT chunk does the same. The only multi-word particle is {\em on and off}. This chunk type should be easy to recognize as it should coincide with the part-of-speech tag RP, but through tagging errors it is sometimes also assigned IN (preposition) or RB (adverb).

INTJ is an interjection phrase/chunk like {\em no, oh, hello, alas, good grief!}. It is quite rare.

The list marker LST is even rarer. Examples are {\em 1., 2., 3., first, second, a, b, c}. It might consist of two words: the number and the period.

The UCP chunk is reminiscent of the UCP (unlike coordinated phrase) constituent in the Treebank. Arguably, the conjunction is the head of the UCP, so most UCP chunks consist of conjunctions like {\em and} and {\em or}. UCPs are the rarest chunks and are probably not very useful for other NLP tasks.

\subsection{Tokens outside}
Tokens outside any chunk are mostly punctuation signs and the conjunctions in ordinary coordinated phrases. The word {\em not} may also be outside of any chunk. This happens in two cases: Either {\em not} is not inside the VP constituent in the Treebank annotation e.g. in
\begin{quote}
... (VP have (VP told (NP-1 clients) (S (NP-SBJ *-1) not (VP to (VP ship (NP anything))))))
\end{quote}
or {\em not} is not followed by another verb (because the main verb is a form of {\em to be}). As the right chunk boundary is defined by the chunk's head, i.e. the main verb in this case, {\em not} is then in fact a postmodifier and as such not included in the chunk: ``... $[_{\rm SBAR}$~that~] $[_{\rm NP}$~there~] $[_{\rm VP}$~were~] n't $[_{\rm NP}$~any major problems~] .''

\subsection{Problems}
All chunks were automatically extracted from the parsed version of the Treebank, guided by the tree structure, the syntactic constituent labels, the part-of-speech tags and by knowledge about which tags can be heads of which constituents. However, some trees are very complex and some annotations are inconsistent. What to think about a VP in which the main verb is tagged as NN (common noun)? Either we allow NNs as heads of VPs (not very elegant but which is what we did) or we have a VP without a head. The first solution might also introduce errors elsewhere... As Ramshaw and Marcus \shortcite{ramshaw95} already noted: ``While this automatic derivation process introduced a small percentage of errors on its own, it was the only practical way both to provide the amount of training data required and to allow for fully-automatic testing.''

\section{Data and Evaluation}

For the CoNLL shared task, we have chosen to work with the same
sections of the Penn Treebank as the widely used data set for 
base noun phrase recognition \cite{ramshaw95}:
WSJ sections 15--18 of the Penn Treebank as training material and 
section 20 as test material\footnote{ 
 The text chunking data set is available at
 http://lcg-www.uia.ac.be/conll2000/chunking/
}.
The chunks in the data were selected to match the descriptions in the
previous section.
An overview of the chunk types in the training data can be found in
table \ref{tab-types}.
De data sets contain tokens (words and punctuation marks), information 
about the location of sentence boundaries and information about chunk 
boundaries. 
Additionally, a part-of-speech (POS) tag was assigned to each token by a
standard POS tagger (Brill \shortcite{brill94} trained on the Penn 
Treebank).
We used these POS tags rather than the Treebank ones in order to make
sure that the performance rates obtained for this data are realistic
estimates for data for which no treebank POS tags are available.

\begin{table}[t]
\begin{center}
\begin{small}
\begin{tabular}{|r|r|l|}\hline
count & \%   & type\\\hline
55081 & 51\% & NP      (noun phrase)\\
21467 & 20\% & VP      (verb phrase)\\
21281 & 20\% & PP      (prepositional phrase)\\
 4227 &  4\% & ADVP    (adverb phrase)\\
 2207 &  2\% & SBAR    (subordinated clause)\\
 2060 &  2\% & ADJP    (adjective phrase)\\
  556 &  1\% & PRT     (particles)\\
   56 &  0\% & CONJP   (conjunction phrase)\\
   31 &  0\% & INTJ    (interjection)\\
   10 &  0\% & LST     (list marker)\\
    2 &  0\% & UCP     (unlike coordinated phrase)\\\hline
\end{tabular}
\end{small}
\end{center}
\caption{
Number of chunks per phrase type in the training data (211727 tokens,
106978 chunks).
}
\label{tab-types}
\end{table}

In our example sentence in section \ref{sec-task}, we have used
brackets for encoding text chunks.
In the data sets we have represented chunks with three types of tags:

\begin{quote}
\begin{tabular}{ll}
B-X & first word of a chunk of type X\\
I-X & non-initial word in an X chunk\\
O   & word outside of any chunk
\end{tabular}
\end{quote}

\noindent
This representation type is based on a representation proposed by
Ramshaw and Marcus \shortcite{ramshaw95} for noun phrase chunks.
The three tag groups are sufficient for encoding the chunks in the
data since these are non-overlapping.
Using these chunk tags makes it possible to approach the chunking
task as a word classification task.
We can use chunk tags for representing our example sentence in the
following way:

\begin{quote}
He/B-NP
reckons/B-VP
the/B-NP
\\
current/I-NP 
account/I-NP 
\\
deficit/I-NP
will/B-VP narrow/I-VP
\\
to/B-PP
only/B-NP $\pounds$/I-NP 
\\
1.8/I-NP billion/B-NP
in/B-PP
\\
September/B-NP
./O
\end{quote}

\noindent
The output of a chunk recognizer may contain inconsistencies in 
the chunk tags in case a word tagged I-X follows a word tagged
O or I-Y, with X and Y being different.
These inconsistencies can be resolved by assuming that such I-X
tags start a new chunk.

The performance on this task is measured with three rates.
First, the percentage of detected phrases that are correct
(precision). 
Second, the percentage of phrases in the data that were
found by the chunker (recall).
And third, the F$_{\beta=1}$ rate which is equal to
($\beta^2$+1)*precision*recall / ($\beta^2$*precision+recall)
with $\beta$=1
\cite{vanrijsbergen75}.
The latter rate has been used as the target for 
optimization\footnote{In the literature about related tasks
sometimes the tagging accuracy is mentioned as well.
However, since the relation between tag accuracy and chunk 
precision and recall is not very strict, tagging accuracy is 
not a good evaluation measure for this task.}.

\begin{table*}[t]
\begin{center}
\begin{tabular}{|l|c|c|c|}\cline{2-4}
\multicolumn{1}{l|}{test data}
                           & precision & recall & F$_{\beta=1}$ \\\hline
Kudoh and Matsumoto        & 93.45\% & 93.51\% & 93.48 \\
Van Halteren               & 93.13\% & 93.51\% & 93.32 \\
Tjong Kim Sang             & 94.04\% & 91.00\% & 92.50 \\
Zhou, Tey and Su           & 91.99\% & 92.25\% & 92.12 \\
D\'ejean                   & 91.87\% & 91.31\% & 92.09 \\
Koeling                    & 92.08\% & 91.86\% & 91.97 \\
Osborne                    & 91.65\% & 92.23\% & 91.94 \\
Veenstra and Van den Bosch & 91.05\% & 92.03\% & 91.54 \\
Pla, Molina and Prieto     & 90.63\% & 89.65\% & 90.14 \\
Johansson                  & 86.24\% & 88.25\% & 87.23 \\
Vilain and Day             & 88.82\% & 82.91\% & 85.76 \\\hline
baseline                   & 72.58\% & 82.14\% & 77.07 \\\hline
\end{tabular}
\end{center}
\caption{Performance of the 
eleven 
systems on the test data.
The baseline results have been obtained by selecting the most frequent
chunk tag for each part-of-speech tag.
} 
\label{tab-results}
\end{table*}

\section{Results}

The 
eleven 
systems that have been applied to the CoNLL-2000 shared task 
can be divided in four groups:

\begin{enumerate}
\itemsep=-0.0cm
\item Rule-based systems:   Villain and Day; 
                            Johansson; 
                            D\'ejean.
\item Memory-based systems: Veenstra and Van den Bosch.
\item Statistical systems:  Pla, Molina and Prieto; Osborne; 
                            Koeling; Zhou, Tey and Su.
\item Combined systems:     Tjong Kim Sang; 
                            Van Halteren;
                            Kudoh and Matsumoto.
\end{enumerate}

\noindent
Vilain and Day \shortcite{vilain2000} approached the shared task in
three different ways.
The most successful was an application of the Alembic parser
which uses transformation-based rules.
Johansson \shortcite{johansson2000} uses context-sensitive and
context-free rules for transforming part-of-speech (POS) tag sequences 
to chunk tag sequences.
D\'ejean \shortcite{dejean2000} has applied the theory refinement
system ALLiS to the shared task.
In order to obtain a system which could process XML formatted data
while using context information, he has used three extra tools.
Veenstra and Van den Bosch \shortcite{veenstra2000} examined different
parameter settings of a memory-based learning algorithm.
They found that modified value difference metric applied to POS
information only worked best.

A large number of the systems applied to the CoNLL-2000 shared task
uses statistical methods.
Pla, Molina and Prieto \shortcite{pla2000} use a finite-state version
of Markov Models.
They started with using POS information only and obtained a better
performance when lexical information was used.
Zhou, Tey and Su \shortcite{zhou2000} implemented a chunk tagger based
on HMMs. 
The initial performance of the tagger was improved by a post-process
correction method based on error driven learning and by incorporating 
chunk probabilities generated by a memory-based learning process.
The two other statistical systems use maximum-entropy based methods.
Osborne \shortcite{osborne2000} trained Ratnaparkhi's maximum-entropy
POS tagger to output chunk tags.
Koeling \shortcite{koeling2000} used a standard maximum-entropy 
learner for generating chunk tags from words and POS tags.
Both have tested different feature combinations before finding an 
optimal one and their final results are close to each other.

Three systems use system combination.
Tjong Kim Sang \shortcite{tks2000} trained and tested five memory-based
learning systems to produce different representations of the chunk tags.
A combination of the five by majority voting performed better than the 
individual parts.
Van Halteren \shortcite{hvh2000} used Weighted Probability
Distribution Voting (WPDV) for combining the results of four WPDV
chunk taggers and a memory-based chunk tagger.
Again the combination outperformed the individual systems.
Kudoh and Matsumoto \shortcite{kudoh2000} created 231 support vector
machine classifiers to predict the unique pairs of chunk tags.
The results of the classifiers were combined by a dynamic programming
algorithm.

The performance of the systems can be found in Table \ref{tab-results}.
A baseline performance was obtained by selecting the chunk tag most
frequently associated with a POS tag.
All systems outperform the baseline.
The majority of the systems reached an F$_{\beta=1}$ score between 
91.50 and 92.50.
Two approaches performed a lot better: the combination system WPDV
used by Van Halteren and the Support Vector Machines used by Kudoh and
Matsumoto.

\section{Related Work}

In the early nineties, Abney \shortcite{abney91} proposed to approach 
parsing by starting with finding related chunks of words.
By then, Church \shortcite{church88} had already reported on
recognition of base noun phrases with statistical methods. 
Ramshaw and Marcus \shortcite{ramshaw95} approached chunking by using
a machine learning method. 
Their work has inspired many others to study the application of 
learning methods to noun phrase chunking\footnote{An elaborate 
overview of the work done on noun phrase chunking can be found on
http://lcg-www.uia. ac.be/\~{ }erikt/research/np-chunking.html}. 
Other chunk types have not received the same attention as NP chunks. 
The most complete work is Buchholz et al. \shortcite{buchholz99},
which presents results for NP, VP, PP, ADJP and ADVP chunks. 
Veenstra \shortcite{veenstra99} works with NP, VP and PP chunks. 
Both he and Buchholz et al. use data generated by the script that 
produced the CoNLL-2000 shared task data sets.
Ratnaparkhi \shortcite{ratnaparkhi98} has recognized arbitrary chunks 
as part of a parsing task but did not report on the chunking 
performance.
Part of the Sparkle project has concentrated on finding various sorts 
of chunks for the different languages \cite{carroll97}.

\section{Concluding Remarks}

We have presented an introduction to the CoNLL-2000 shared task:
dividing text into syntactically related non-overlapping groups of
words, so-called text chunking.
For this task we have generated training and test data from the Penn
Treebank.
This data has been processed by eleven systems.
The best performing system was a combination of Support Vector
Machines submitted by Taku Kudoh and Yuji Matsumoto.
It obtained an F$_{\beta=1}$ score of 93.48 on this task.

\section*{Acknowledgements}

We would like to thank
the members of the CNTS - Language Technology Group in Antwerp,
Belgium and the members of the ILK group in Tilburg, The Netherlands 
for valuable discussions and comments.
Tjong Kim Sang is funded by the European TMR network Learning
Computational Grammars. 
Buchholz is supported by the Netherlands Organization for Scientific
Research (NWO).

\small
\bibliographystyle{acl}

\end{document}